\documentclass[conference]{IEEEtran}
\IEEEoverridecommandlockouts
\usepackage{cite}
\usepackage{amsmath,amssymb,amsfonts}
\usepackage{algorithmic}
\usepackage{graphicx}
\usepackage{textcomp}
\usepackage{xcolor}

\usepackage{booktabs}
\usepackage[colorlinks=true, allcolors=blue]{hyperref}
\usepackage{float} 
\usepackage{subfigure} 
\usepackage{multicol}

\def\BibTeX{{\rm B\kern-.05em{\sc i\kern-.025em b}\kern-.08em
    T\kern-.1667em\lower.7ex\hbox{E}\kern-.125emX}}
\begin{document}

\title{Spatio-Temporal Similarity Measure based Multi-Task Learning for Predicting Alzheimer’s Disease Progression using MRI Data}

\author{
\IEEEauthorblockN{Xulong Wang}
\IEEEauthorblockA{\textit{Department of Computer Science} \\
\textit{University of Sheffield}\\
Sheffield, UK \\
xl.wang@sheffield.ac.uk}
\and
\IEEEauthorblockN{Yu Zhang}
\IEEEauthorblockA{\textit{Department of Computer Science} \\
\textit{University of Sheffield}\\
Sheffield, UK \\
yzhang489@sheffield.ac.uk}
\and
\IEEEauthorblockN{Menghui Zhou}
\IEEEauthorblockA{\textit{Department of Computer Science} \\
\textit{University of Sheffield}\\
Sheffield, UK \\
mzhou47@sheffield.ac.uk}
\and
\IEEEauthorblockN{Tong Liu}
\IEEEauthorblockA{\textit{Department of Computer Science} \\
\textit{University of Sheffield}\\
Sheffield, UK \\
tliu.soton@gmail.com}
\and
\IEEEauthorblockN{Jun Qi}
\IEEEauthorblockA{\textit{Department of Computing} \\
\textit{Xi’an JiaoTong-Liverpool University}\\
Suzhou, China \\
Jun.Qi@xjtlu.edu.cn}
\and
\IEEEauthorblockN{Po Yang$^*$ \thanks{\IEEEauthorrefmark{4} Po Yang is the corresponding author.}}
\IEEEauthorblockA{\textit{Department of Computer Science} \\
\textit{University of Sheffield}\\
Sheffield, UK \\
po.yang@sheffield.ac.uk}
}

\maketitle

\begin{abstract}
Identifying and utilising various biomarkers for tracking Alzheimer's disease (AD) progression have received many recent attentions and enable helping clinicians make the prompt decisions. Traditional progression models focus on extracting morphological biomarkers in regions of interest (ROIs) from MRI/PET images, such as regional average cortical thickness and regional volume. They are effective but ignore the relationships between brain ROIs over time, which would lead to synergistic deterioration. For exploring the synergistic deteriorating relationship between these biomarkers, in this paper, we propose a novel spatio-temporal similarity measure based multi-task learning approach for effectively predicting AD progression and sensitively capturing the critical relationships between biomarkers. Specifically, we firstly define a temporal measure for estimating the magnitude and velocity of biomarker change over time, which indicate a changing trend(temporal). Converting this trend into the vector, we then compare this variability between biomarkers in a unified vector space(spatial). The experimental results show that compared with directly ROI based learning, our proposed method is more effective in predicting disease progression. Our method also enables performing longitudinal stability selection to identify the changing relationships between biomarkers, which play a key role in disease progression. We prove that the synergistic deteriorating biomarkers between cortical volumes or surface areas have a significant effect on the cognitive prediction.
\end{abstract}


\section{Introduction}
Alzheimer's disease (AD) is a serious neurodegenerative disease, which is characterized by memory loss and cognitive decline due to the progressive damage of neurons and their connections, which directly leads to death \cite{alzheimer20192019}. According to World Health Organization (WHO), it is estimated that there are globally 47.5 million people with dementia in 2016 with 7.7 million new cases every year.
\begin{figure*}[!t] 
\centering 
\includegraphics[width=1\textwidth]{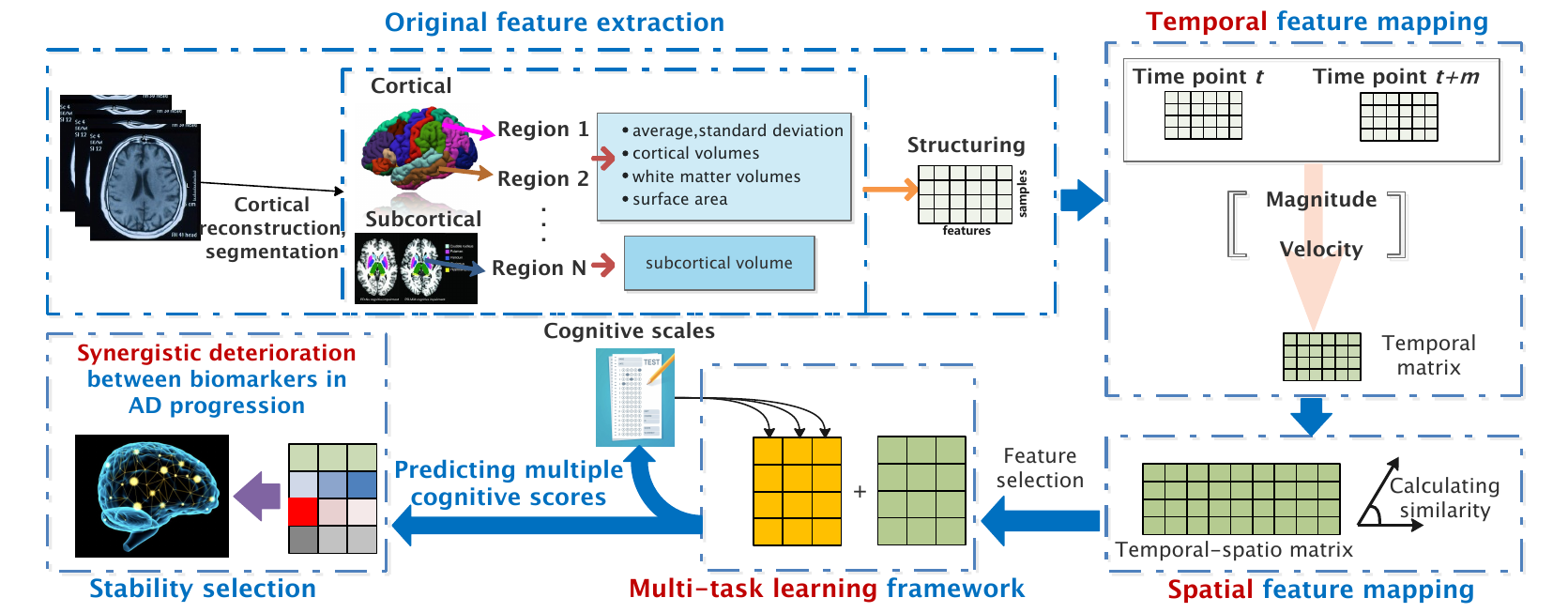} 
\caption{The protocol of MTL approach using spatio-temporal measure.} 
\label{Fig.1} 
\end{figure*}
Previous research has focused on using biomarkers combined with machine learning algorithms to predict patients' Mini Mental State Examination (MMSE) and Alzheimer's Disease Assessment Scale cognitive subscale (ADAS-Cog) scores as the target data to predict whether a patient is an AD patient and find the weight of each biomarker feature at different prediction time points, existing AD disease progression models mainly use machine learning regression algorithms \cite{tabarestani2020distributed}, survival models based on statistical probabilities \cite{doody2010predicting,green2011model}, and deep learning methods based on neural networks \cite{nguyen2018modeling,liu2014early,yang2019duapm}. The above-mentioned research focuses on using the data obtained by the patient during the first test (baseline data) to make predictions, which is a method that uses a small number of input features to make predictions. The disadvantage is that it ignores the information contained in the biomarkers in the process of changing over time.

Previous studies focusing ROIs of brain have studied the differences in the correlation between brain biomarkers for AD, cognitively normal older individuals (NL) and mild cognitive impairment (MCI).
\cite{sivera2019model} proposed a deformation-based framework to jointly model the effects of aging and AD on the evolution of brain morphology, confirming the existence of components that significantly accelerate aging in AD patients. \cite{vemuri2009mri} evaluated the correlation of MRI and CSF biomarkers with clinical diagnosis and cognitive performance in subjects with NL and aMCI (amnestic mild cognitive impairment) and AD patients. 
It is concluded that MRI provides stronger cross-sectional grouping and recognition ability and has better correlation with general cognitive and functional status on the cross-section, and MRI can reflect the clinically determined disease stage than CSF biomarkers.  
On the longitudinal studies, \cite{planche2022structural} described a novel perspective on volume trajectories and brain atrophy progression of single biomarkers' differences between normal aging and AD.
Some previous studies focused on the similarity between biomarkers from ROI, \cite{wee2013prediction} employed the correlation of regional average cortical thickness and multi-kernel support vector machine to integrate relevant information with ROI-based information to improve the classification performance. 
However, the above-mentioned researches only focused on the use of a single biomarker or the same type of biomarkers and did not focus on the relationships of temporal and spatial changes between different types of biomarkers.

To address the above challenges and uncover the critical relationships between biomarkers, we propose to utilise the temporal and spatial information of brain changes to model the disease process of AD. 
Additionally, to reinforce temporal relationships between follow-up time points, a multi-task learning method\cite{zhou2011malsar} based on temporal smoothness is introduced for interpretably modelling disease progression.


In this paper, we propose to utilise the spatio-temporal similarity between biomarkers changes to predict clinical scores of patients. Specifically, we firstly define a temporal measure for estimating the magnitude and velocity of biomarker
change over time, which indicate a changing trend(Fig.\ref{Fig.1}:temporal feature mapping).
Converting this trend into the vector, we then compare this variability between biomarkers in a unified vector space(Fig.\ref{Fig.1}:spatial feature mapping).
The computation of spatial similarity results in an increase in data dimension by an order of magnitude of square.
Faced with the scarcity of samples and a large number of feature dimensions, we introduce multiple loss terms with  $\mathcal{L}_1$ \cite{tibshirani1996regression} and its variant norm \cite{zhou2011malsar} to overcome the \href{https://en.wikipedia.org/wiki/Curse_of_dimensionality}{Curse of dimensionality} and interpretably capture the key relationships.
The contributions of this work are summarized as follows:
\begin{itemize}
    \item A novel spatio-temporal similarity measure approach is proposed of analysing and extracting reliable features from MRI. This similarity measure will effectively quantify the synergistic deterioration between these biomarkers over time;
    \item A multi-task learning (MTL) algorithm with spatio-temporal embedding is designed for effectively predicting AD progression, visualising brain biomarkers related to this progression;
    \item A comprehensive experimental analysis is carried out by accessing impact of AD progression on brain function synergistic deteriorating biomarkers.
\end{itemize}

\section{Related work}
In traditional machine learning paradigm, an accurate learner is usually treated as one single learning task (e.g., classification, regression) and learnt by a large number of training samples.
For instance, deep learning model \cite{nguyen2018modeling,ghazi2019training} can train an accurate AD prediction model of neural network with hundreds of layers contacting a great amount of parameters via massive labelled biomarkers at baseline from ADNI. But one key challenge here is that sufficient and well-labelled longitudinal AD data at multiple time points are hardly collected from AD patients. The problem of missing, sparse and insufficient data strongly impacts on learning a fine model. Differing with traditional ML approaches, Multi-Task Learning \cite{thung2018brief} considers the prediction of AD progression as multiple learning tasks each of which can be a general prediction task art certain time point. Among these prediction tasks, all of them are assumed to be related to each other in time domain with relevant temporal features (e.g., biomarkers in MRI). We demonstrate a typical pipeline of leveraging MTL algorithms for predicting cognitive functionality of AD patients from their brain imaging scans \cite{zhou2011multi}, where the predictive information is shared and transferred among related models to reinforce their generalization performance. The data sources employed are Freesufer (Extracted features from MRI like Volume of Hippocampus) and cognitive functional scores (AD cognitive scales like MMSE\cite{kurlowicz1999mini} or ADAS-cog \cite{chu2000reliability}) from selected AD patients repeatedly by multiple time points. By considering the prediction of cognitive scales at a single time point (like 6, 12 or 18 months) as a regression task. The prediction of clinical scores at multiple future time points as a multi-task regression problem. Weights of MTL are trained and optimized through processing pre-extracted features from MRI and baseline cognitive scales.

Two important issues affect the progress of applying MTL in AD modelling problems. First, it is important to obtain good quality of baselines from AD raw data, where MRI reflects changes in brain structure, such as the cerebral cortex and ventricle; cognitive scale directly shows cognitive functions of AD patients. Sparse representation \cite{qi2020overview} is a popular method in MTL for capturing key biomarkers in AD, which uses sparseness as a regularization condition, image blocks with key characteristics. Cognitive measure can be achieved by using worldwide standard AD cognitive assessment, such as MMSE \cite{kurlowicz1999mini}, ADAS-cog \cite{chu2000reliability} and Rey Auditory Verbal Learning Test (RAVLT)\cite{schmidt1996rey,vakil1993rey}. As the second issue, utilizing and improving advanced regression models \cite{wan2014identifying} in MTL are highly critical, where they could better explore the relationship and correlations between MRI features and cognitive measures. Here, structural regularization \cite{zhou2011malsar} is a common approach in MTL for minimize the penalized empirical loss and bundling the correlations between tasks in the assumption. In the field of MTL in AD, there are many prior work that model relationships among tasks using novel regularizations \cite{cao2017sparse,wang2019multi}. The addition of kernel method problems allows the algorithm to fit non-linear relationships \cite{peng2019structured}. The benchmark of this paradigm is derived from \cite{zhou2013modeling} and subsequent achievements are mostly aimed at theoretical structure, relevance, and fusing the multi-modality data applications. So far to our best knowledge, above regularized MTL approaches deliver promising performance in many AD prediction applications.

\section{Methodology}
\subsection{Problem formulation}
Consider a MTL of $k$ tasks with $n$ training samples of $d$ features. Let ${x_1,x_2,..., x_n}$be the input data for the patients, and ${y_1,y_2,...,y_n}$be the predicted cognitive scale for each patient, where each $x_i \in \mathbb{R}^d$ represents the feature data of an AD patient, and $y_i \in \mathbb{R}$ is the predicted value of different cognitive scales. Specifically, $x_i^j = [{\mathrm{m}},{\mathrm{v}}] $ denotes spatio-temporal ROIs similarity on features $j_{th} $ and $ (j+r)_{th}$  of the $i_{th}$ sample, $\mathrm{m},{\mathrm{v}}$ represent the magnitude and velocity of two specific biomarkers over time scale, where $j,(j+r) \in (0,d]$.

Then, let $ X = [x_1, ..., x_n]^T \in \mathbb{R}^{n\times d}$ be the data matrix, $Y = [y_1,..., y_n]^T \in \mathbb{R}^{n \times k} $ be the predicted matrix, and $ W = [w_1, ..., w_k]^T \in \mathbb{R}^{d\times k}$ be the weight matrix. The process of establishing a MTL model is to estimate the value of W, which is the parameter to be estimated from the training samples. 

In order to solve above problem, many prior works in MTL that model relationships among tasks using regularization methods. Normally, they assume the empirical loss to be square loss and common regularization terms are $\mathcal{L}_1$ and $\mathcal{L}_2$ norms, separately named as Lasso regression and ridge regression models as shown in Eq. \ref{lasso} and \ref{ridge}. Ridge regression constrains variables to a smaller range for reducing some factors with little impacts on model’s prediction. Unfortunately, this reduction means that these variables are still considered. To solve this problem, Lasso was proposed as a new sparse representation linear algorithm, which simultaneously performs feature selection and regression. Some variables are set to zero directly to achieve sparsity and dimensionality reduction.
\begin{equation}
    \min_{w} L(Y,X,W) + \lambda ||W||_1  \label{lasso}
\end{equation}
\begin{equation}
    \min_{w} L(Y,X,W) + \lambda ||W||_2 \label{ridge}
\end{equation}

In AD study, the task of predicting AD patient’s cognitive scale at certain time point is strongly associated with other tasks at adjacent time points. Thus, many recent studies have focused on designing novel structural regularization methods to improve their performance in AD study.

In this paper, we concentrate on two AD progression prediction models : Temporal Group Lasso (TGL)\cite{zhou2011multi} and Convex Fused Sparse Group Lasso (cFSGL)\cite{zhou2012modeling}. Specifically, TGL contains a time smoothing term and a group Lasso term as constraints, which ensures that all regression models at different time points share a common set of features. The TGL formulation solves the following convex optimization problem:
\begin{equation}
    \min_w ||XW - Y||^2_F + \theta_1||W||^2_F + \theta_2||WH||^2_F + \delta||W||_{2,1} \label{tgl}
\end{equation}
where the first term measures the empirical error on the training data, $||W||_F$ is the Frobenius norm, $||WH||^2_F$ is the temporal smoothness term, which ensures a small deviation between two regression models at successive time points, and $||W||_{2,1}$ is the group lasso penalty, which ensures that a small subset of features will be selected for the regression models at all-time points.

cFSGL involves sparsity between tasks, where it considers both common features at different points in time and unique features to each task. This feature is helpful to improve the overall performance of the model. cFSGL formulation solves the following convex optimization problem:
\begin{equation}
    \min_w ||XW - Y||^2_F + \theta_1||W||_1 + \theta_2||RW^T||_1 + \delta||W||_{2,1} \label{cfsgl}
\end{equation}
where the first term measures the empirical error on the training data, $||W||_1$ is the lasso penalty, $||RW^T||_1$ is the fused lasso penalty, and $||W||_{2,1}$ is the group lasso penalty.

Lasso and group lasso combined employ is called sparse group lasso, which allows simultaneous selection of a common feature for all time points and internally generates sparse solutions in response to different time points. Fused lasso penalty having a given temporal smoothness, which makes selected features at nearby time points similar to each other. In addition, notice that cFSGL's formula involves three non- smooth terms. Accelerated gradient descent method is utilised to solve this problem.

\subsection{Definition of spatio-temporal similarity}
Two consecutive MRI scans are used to calculate the temporal and spatial changes of brain biomarkers. For instance, we utilise BL and M06 MRI to calculate the magnitude and velocity for biomarkers, let x be the detection value of brain biomarkers and t be the MRI test dates, the magnitude is $\frac{x_{M06}-x_{BL}}{x_{BL}}$, the velocity is $\frac{x_{M06}-x_{BL}}{t_{M06}-t_{BL}}$ per month. Use the magnitude and velocity to compose a vector that represents the changing trend of the brain biomarker.

Cosine similarity is used to calculate the similarity between two vectors to express the similarity of the temporal and spatial changes of two MRI biomarkers. Cosine similarity uses the cosine value of the angle between two vectors in the vector space as a measure of the difference between two individuals. As the values of different types of biomarkers are different in MRI dataset, while the cosine similarity measures the difference in trend rather than the value. The temporal and spatial relationships of brain biomarkers of AD, NL and MCI displayed by cosine similarity, euclidean distance and mahalanobis distance.

\subsection{Experiment protocol}
Firstly, we verified that MTL is superior in following AD progression.
Combined with randomization techniques, we locate stable and sensitive cortical biomarkers identified by MTL algorithm. 
Our empirical protocol design are shown in Fig.\ref{Fig.1}. 
The complete experimental process mainly includes 6 steps:
\begin{enumerate}
    \item \textit{Original feature extraction.} Statistical features based on ROIs in cerebral cortex/sub-cortex are extracted from MRI images.
    \item \textit{Temporal feature mapping.} Potential biomarkers (ROIs) change over time are characterized from the magnitude and velocity. Describing temporal changes in biomarkers using a two-dimensional vector.
    \item \textit{Spatial feature mapping.} Calculating spatial similarity (cosine similarity) between vectors. 
    \item \textit{Feature selection.} Through this stage, the features dimension is greatly reduced, and the key features of temporal-spatial is retained. 
    \item \textit{Predicting multiple cognitive scores.} Modelling the AD progression between biomarkers and cognitive scales via MTL methods.
    \item \textit{Stability selection.} Embedding MTL methods in the general stability selection to excavate synergistic deterioration between biomarkers in AD progression.
\end{enumerate}

Secondly, cross-validation is employed to split the training and test data. 
We utilise different metrics to evaluate the model performance on test data.
The regression performance metric often employed in MTL is normalized mean square error (nMSE) and root mean square error (rMSE) is employed to measure the performance of each specific regression task. In particular, nMSE has been normalized to each task before evaluation, so it is widely used in MTL methods based on regression tasks. Also, weighted correlation coefficient (wR) as employed in the medical literature addressing AD progression problems\cite{zhou2013modeling,ito2011disease,stonnington2010predicting}. 
nMSE, rMSE and wR are defined as follows:
\begin{align}
& \operatorname{nMSE}(Y, \hat{Y})=\frac{\sum_{i=1}^t\left\|Y_i-, \hat{Y}_i\right\|_2^2 / \sigma\left(Y_i\right)}{\sum_{i=1}^t n_i} \\
& \operatorname{rMSE}(y, \hat{y})=\sqrt{\frac{\|y-\hat{y}\|_2^2}{n}} \\
& \operatorname{wR}(\mathrm{Y}, \hat{Y})=\frac{\sum_{i=1}^t \operatorname{Corr}\left(\mathrm{Y}_i, \hat{Y}_i\right) n_i}{\sum_{i=1}^t n_i}
\end{align}

Finally, as for repeated experimental times, one evaluation consensus in MTL models for AD study is that one experiment result is usually accidental and unreliable. To reduce experiment accidental errors, repeated experiments are required. We also evaluate the performance of four selected regularized MTL models under different repeated experimental times and lastly evaluate typical factors like data size and number of tasks affecting MTL models.

\subsection{Stability Selection via MTL}
In order to improve the interpretability and robustness of the results, stability selection was modified to meet our actual needs. The original strategy of feature selection was included a Lasso algorithm as core feature subsets searches approaches. In this paper, MTL algorithms were utilised to embedded in stability selection.

Let $F$ be the overall set of features and let $f \in F$ be the subset of features by sub-sampling. Let $\gamma$  denote the iteration number of sub-sampling and $D{i}=\{ X(i), Y(i) \}$ denote one random sub-sample operation of number $i \in ( 0, \gamma ] $. Each operation size account for $\llcorner \frac{n}{2} \lrcorner$. Let $\Lambda $ be the regularization parameter space. For a $\lambda \in \Lambda$, let $\hat{W}^{(i)}$ denote the model coefficient of MTFL that fitted on a subset of $D(i)$. Then, the subset of features generated in task $j$ by the sparse constraints of the MTFL algorithm can be denote as:
\begin{equation}
S_j^\lambda\left(D_{(i)}\right)=\left\{f: \hat{W}_j^{(i)} \neq 0\right\} . \label{sd}
\end{equation}
With stability selection, we do not simply select one model in the parameter space $\lambda$. Instead the data are perturbed (e.g. by sub-sampling) $\gamma$ times at task j and we choose all structures or variables that occur in a large fraction of the resulting selection sets:
\begin{equation}
\hat{\pi}_j^\lambda=\frac{\sum_{i=1}^\gamma I\left(f \in S_j^\lambda\left(D_{i j}\right)\right)}{\gamma} . \label{pi}
\end{equation}
Where indicator function $I(\bullet)$ denote 
$I(x) =
\begin{cases} 
1,  & x=0 \\
0, & others
\end{cases}$
and $\hat{\pi}_j^\lambda \in [0,1]$ denote the stability probability of task $j$ at MTFL approaches which feature selection is not based on individual operations but on multiple task collaboration constraints.

Repeat the above procedure for all $\lambda \in \Lambda$, we obtain the stability score $S_j(f)$ for each feature $f$ at task $j$:
\begin{equation}
S_j(f)=\max _{\lambda \in \Lambda}\left(\hat{\pi}_j^\lambda\right) . \label{sf}
\end{equation}

Finally, for a cut‐off $\pi_{th}$ with $0<\pi_{th}<1$ and a set of regularization parameters $\Lambda$, the set of stable variables is defined as:
\begin{equation}
\hat{S}^{\text {stable }}=\left\{k: S_j(f) \geq \pi_{t h}\right\}=\left\{k: \max _{\lambda \in \Lambda}\left(\hat{\pi}_j^\lambda\right) \geq \pi_{th}\right\} . \label{sstable}
\end{equation}
The embedded multi-task approach ensures that the selected features have the following properties:1) Stability. A cortical region of the brain that is closely related to the subject's disease progression. 2) Global significance. MTL makes sure that the selected features are important for each task. One technique that arises here is to pick the coefficient value for one of the tasks when doing statistics on the stability of the selected features at equation \ref{cfsgl}.

\section{experimental settings}
\subsection{Subjects}

\begin{table}[]
\centering
\caption{SCREENING SUBJECTS}
\label{subject}
\resizebox{\columnwidth}{!}{%
\begin{tabular}{@{}cccc@{}}
\toprule
Time Span         & Scanning Subjects & MMSE & Baseline Subjects \\ \midrule
Baseline   to M06 & 700               & 429  & 408               \\
Baseline   to M12 & 670               & 429  & 402               \\
Baseline   to M24 & 533               & 429  & 373               \\
Baseline   to M36 & 337               & 429  & 327               \\ \bottomrule
\end{tabular}%
}
\end{table}


To track the effectiveness of disease progression models,
\href{https://adni.loni.usc.edu/about/adni1/}{ADNI-1} subjects with all corresponding MRI and cognitive scales are evaluated.  As shown in the Table \ref{subject}.
Subjects are between 55–90 years of age, the male accounts for 52.18\%, the degree of suffering from the dementia, the data ratio of AD, MCI and NL are 25\%, 50\% and 25\% respectively.

To explore the impact of the correlation between ROIs on AD progression, MRI data from two follow-up points in the longitudinal cohort were extracted to facilitate observation of this spatiotemporal variation. At the same time, the cognitive scales (like MMSE or ADAS-cog) of longitudinal cohorts are employed to estimate the patients’ cognitive functional decline during the AD progression. During the screening period, all the subject must satisfy the data integrity for verifying the reliable result. Namely, the cohort subjects must complete participation in two follow-up point MRI scans and multiple cognitive scoring assessments.

\subsection{Data pre-processing}
For guarantees high image quality and reliable data handling, the MR images used in the paper were derived from standardized datasets, which provide the intensity normalized and gradient unwrapped TI image volumes. Subsequently, the FreeSurfer\cite{fischl2012freesurfer} was performed to feature extraction of the MR, which execute cortical reconstruction and volumetric segmentations for processing and analysing brain MR images.

For each MRI, cortical regions and subcortical regions are generated after this pre-processing suite. For each cortical region, the cortical thickness average, standard deviation of thickness, surface area, and cortical volume were calculated as features. For each subcortical region, subcortical volume was calculated as feature. Data cleaning operations are performed as the following steps: 1) removal of individuals who failed cortical reconstruction and failed quality control; 2) removal of features with more than half of the sample missing values; 3) individual subject whose removal of baseline did not screen for MRI; 4) using the average of the features to fill in missing data; and 5) removal of cognitive function tests in individuals with missing follow-up points in longitudinal studies.

\subsection{Feature selection}
To discover the impact of the similarity between ROIs on progression with AD, we couple all the regions in pairs, which allows 326 ROIs statistic features to combine 52975 features. For a given sample size, the higher the dimensionality, the sparser the distribution of the sample in space. 

To solve this issue, we utilised TGL combined with a stability selection algorithm to obtain features that play an important role for all tasks.
Finally, 300 significant features are selected for the training of MTL algorithm.

\section{Experimental results and analysis}
\subsection{Spatio-Temporal similarity measure}
We first accomplish three relevance approaches of estimating ROIs relevant criteria: Euclidean Distance (ED), Mahalanobis Distance (MD) and Cosine Similarity (CS). And then, each criterion between vectors composed of the magnitude and the non-absolute value of velocity of the biomarker are used and the feature subset selecting the original feature space to evaluate the subjects cognitive scales. Table \ref{Different Similarity measure} shows the different criterion tracking the AD progression. Note that Table \ref{Different Similarity measure} shows only the averaged results and variance of 30 independent experiments; and the temporal distance from baseline to M06 period. Besides, we also reproduced the model achieved by \cite{zhou2011multi,zhou2012modeling,zhou2013modeling}, with only MRI data as features. 

\begin{table}[t]
\centering
\caption{Different Similarity measure}
\label{Different Similarity measure}
\resizebox{\columnwidth}{!}{%
\begin{tabular}{@{}ccccc@{}}
\toprule
                 & Original ROI         & Mahalanobis Distance & Euclidean Distance   & Cosine Similarity    \\ \midrule
\textit{Target: MMSE}     &                      &                      &                      &                      \\
nMSE             & 0.827±0.065          & 0.973±0.076          & 0.944±0.080          & \textbf{0.743±0.060} \\
wR               & 0.461±0.053          & 0.273±0.080          & 0.552±0.043          & \textbf{0.552±0.043} \\
BL rMSE          & 1.750±0.157          & 1.782±1.509          & 1.901±0.176          & \textbf{1.436±0.134} \\
M06 rMSE         & 2.326±0.302          & 2.420±0.195          & 2.240±0.286          & \textbf{2.190±0.215} \\
M12 rMSE         & 2.599±0.366          & 2.943±0.291          & \textbf{2.505±0.415} & 2.541±0.415          \\
M24 rMSE         & 3.516±0.777          & 3.747±0.627          & 3.689±0.694          & \textbf{3.227±0.575} \\
M36 rMSE         & 4.169±0.831          & 4.394±0.803          & 5.020±0.866          & \textbf{4.125±0.846} \\
\textit{Target: ADAS-cog} &                      &                      &                      &                      \\
nMSE             & 0.693±0.054          & 0.773±0.087          & 0.790±0.067          & \textbf{0.666±0.058} \\
wR               & 0.579±0.041          & 0.514±0.057          & 0.488±0.055          & \textbf{0.604±0.043} \\
BL rMSE          & 4.093±0.388          & 3.809±0.371          & 4.238±0.470          & \textbf{3.670±0.606} \\
M06 rMSE         & 4.540±0.609          & 4.375±0.497          & 4.665±0.529          & \textbf{4.399±0.740} \\
M12 rMSE         & 4.932±0.781          & 4.759±0.627          & 4969±0.590           & \textbf{4.693±0.562} \\
M24 rMSE         & \textbf{5.466±0.774} & 6.234±1.104          & 6.537±1.023          & 5.706±0.899          \\
M36 rMSE         & \textbf{7.661±1.092} & 8.943±1.969          & 8.851±1.352          & 8.133±1.720          \\ \bottomrule
\end{tabular}%
}
\end{table}

\begin{table}[t]
\centering
\caption{ROIS SYNCHRONIZATION REPRESENTS THE PROGRESS OF AD}
\label{ROIS SYNCHRONIZATION REPRESENTS THE PROGRESS OF AD}
\resizebox{\columnwidth}{!}{%
\begin{tabular}{@{}llllll@{}}
\toprule
                 & Original ROI         & BL to M06   & BL to M12   & BL to M24            & BL to M36            \\ \midrule
\textit{Target: MMSE}     &                      &             &             &                      &                      \\
nMSE             & 0.827±0.065          & 0.743±0.060 & 0.726±0.092 & \textbf{0.693±0.070} & 0.724±0.121          \\
wR               & 0.461±0.053          & 0.552±0.043 & 0.581±0.047 & \textbf{0.595±0.044} & 0.582±0.065          \\
BL rMSE          & 1.750±0.157          & 1.436±0.134 & 1.472±0.152 & 1.408±0.177          & \textbf{1.335±0.152} \\
M06 rMSE         & 2.326±0.302          & 2.190±0.215 & 2.265±0.268 & 2.134±0.194          & \textbf{1.983±0.332} \\
M12 rMSE         & 2.599±0.366          & 2.541±0.415 & 2.440±0.335 & 2.559±0.481          & \textbf{2.053±0.306} \\
M24 rMSE         & 3.516±0.777          & 3.227±0.575 & 3.197±0.549 & 3.244±0.644          & \textbf{2.710±0.517} \\
M36 rMSE         & 4.169±0.831          & 4.125±0.846 & 4.157±0.704 & 3.847±0.829          & \textbf{3.345±0.701} \\
\textit{Target: ADAS-cog} &                      &             &             &                      &                      \\
nMSE             & 0.693±0.054          & 0.666±0.058 & 0.691±0.087 & \textbf{0.653±0.075} & 0.881±0.056          \\
wR               & 0.579±0.041          & 0.604±0.043 & 0.592±0.051 & \textbf{0.626±0.049} & 0.387±0.053          \\
BL rMSE          & 4.093±0.388          & 3.670±0.606 & 3.522±0.329 & \textbf{3.648±0.479} & 4.084±0.414          \\
M06 rMSE         & 4.540±0.609          & 4.399±0.740 & 4.406±0.514 & \textbf{4.206±0.530} & 4.462±0.609          \\
M12 rMSE         & 4.932±0.781          & 4.693±0.562 & 4.847±0.681 & \textbf{4.872±0.702} & 5.067±0.630          \\
M24 rMSE         & \textbf{5.466±0.774} & 5.706±0.899 & 5.953±0.929 & 5.707±1.124          & 5.530±0.625          \\
M36 rMSE         & \textbf{7.661±1.092} & 8.133±1.720 & 8.100±1.349 & 8.255±1.727          & 7.761±1.518          \\ \bottomrule
\end{tabular}%
}
\end{table}

Overall the cosine similarity representation of our proposed ROIs synchronization approaches outperforms the original ROIs feature. We have the following observations: 1) The collaborative expression of ROIs is better than independent ROI to a certain extent. 2) The expression of cosine similarity performs better than that of cosine similarity and Mahalanobis Distance. 3) The proposed cosine similarity representation witnesses significant improvement for the early time point. This may be due to the data spanning from baseline and M06 period.

\subsection{Modelling AD progression via MTL}
Inspired by the above experiments, we further explored the influence of temporal span on the progress of positioning AD under the collaborative expression of ROIs. In this section, only cosine similarity was utilized to estimate the cognitive functional progression.

There are four temporal span group performed, namely baseline to M06 period, baseline to M12 period, baseline to M24 period and baseline to M36 period. Table \ref{ROIS SYNCHRONIZATION REPRESENTS THE PROGRESS OF AD} shows that the normalized results of different visited time span and the root mean square error of each sub-task results. We follow the same experimental procedure as above. The experimental results are presented in Table \ref{ROIS SYNCHRONIZATION REPRESENTS THE PROGRESS OF AD}.

We can observe from the table that as the time span increases, the overall generalization performance of the model improves. When the temporal span growths, we also have the following observations: 1) The performance of the subtasks will gradually improve. 2) The task of the latter point in time has been greatly enhanced. This may be due to the latter MRI scanning support more collaborative expression of ROIs and these results further validate the efficacy of the proposed method for temporal-spatial collaborative expression of ROIs. 3) during the BL to M24, the overall task performance outperforms others. 4) during the BL to M36, Although the performance of the global model has decreased, the performance of each subtask has been greatly improved. 

\subsection{Stable synergistic deterioration pattern}

\begin{figure}[!t] 
\centering 
\includegraphics[width=0.5\textwidth]{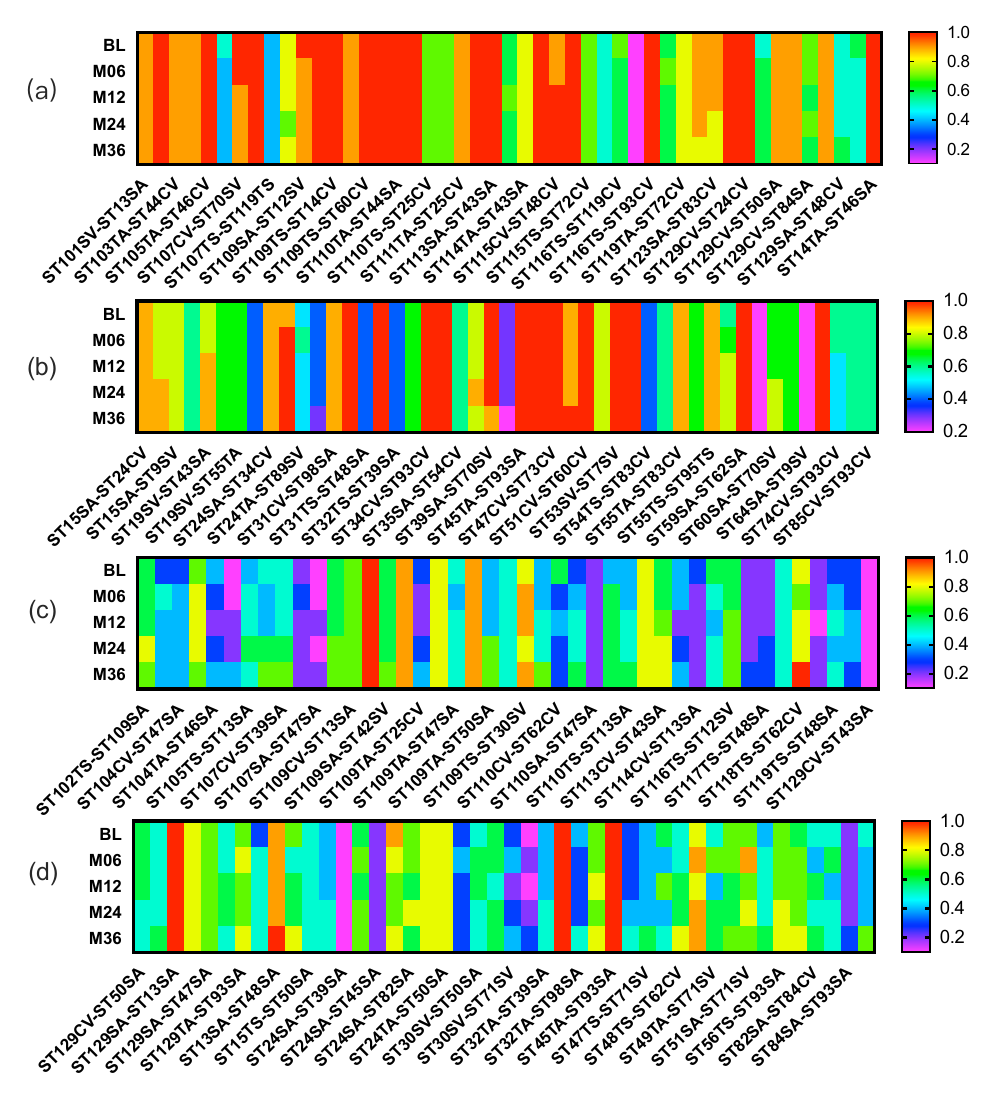} 
\caption{ the vectors of stability temporal collaborative patterns. A total of 94 and 87 stable deteriorating pairs respectively. Specifically, (a) and (b) belong to the MMSE-targeted model of AD progress; (c) and (d) belong to ADAS-cog-targeted model of AD progress.} 
\label{Fig.2} 
\end{figure}

Firstly, we use the data from a set of experiments with the best performance in experiment: Temporal Span of MRI Scan, namely the temporal span for baseline to M24 periods, which contains 94 dimensions corresponding a crucial couples of ROIs pairs. Secondly, a set of environmental parameters are clearly indicated: 1) Only half of the overall sample in each sampling subset is randomly selected. 2) A total of 210 combinations of model hyperparameters. 3) during every combination, 10 samplings were executed. Finally, the vectors of stability temporal collaborative patterns are showed in Fig.\ref{Fig.2}. For the MMSE set, the result shows that the synergistic effect of left insula on left entorhinal cortex, left posterior cingulate cortex, right bankssts, left caudal anterior cingulate cortex, left pars triangularis. The synergistic effect of right posterior cingulate gyrus on right isthmus of cingulate cortex, left temporal pole. For the ADAS-cog set, the result shows that the synergistic effect of left insula on left entorhinal cortex, left posterior cingulate cortex, left bankssts, left pars triangularis. The synergistic effect of left entorhinal on left parahippocampal, right cuneus, medial orbitofrontal cortex. The synergistic effect of right posterior cingulate cortex on left pars triangularis, left parahippocampal. The fact that our findings are in line with those of previous studies\cite{liu2011combination,davatzikos2011prediction,wang2009alterations} demonstrates the validity of our proposed model.

In the selection of longitudinal stability, we observed 29 most stable features with MMSE score, which are shown in Fig.\ref{Fig.2}, where the horizontal axis represents the markers between each salient ROI pair, and details are available in the \nameref{appendix}. 
The correlation features based on Cortical Volume and Cortical Volume are the majority (6 features), which shows that the similarity of the change trend of the biomarkers based on Cortical Volumes have important effect in AD prediction. Previous studies have also observed a significant improvement in the classification performance of abnormal cortical patterns and the coordinated patterns of cortical morphology are widely altered in AD patients \cite{wee2013prediction}. In addition, the number of correlation features based on the similarity of changes between Surface Area and Surface Area is also relatively large (5 features).

\section{Discussion}
Although we modelled the spatio-temporal correlation of ROIs between time points, we focused only on the cognitive scales of the latter time point. 
The alignment of two cognitive scales would provide valuable context in the MTL settings as the cognitive scales might potentially also change over time.
Additionally, we only focuses on the comparison of methods based on temporal smoothness and does not consider methods such as spatial assumptions. More comparisons will be carried out in future work.


\section{conclusion}
Identifying  the synergistic deteriorating relationship of biomarkers can help clinicians assess AD progress in early intervention. We propose a new method to model and predict AD progress by extracting morphological information from MRI. This paper has three main contributions. Firstly, we employ cosine similarity to represent a temporal-spatial relationships between brain biomarkers.
We then regard the disease progression prediction as a MTL problem and combine the cosine similarity to predict the disease progression of AD. 
Finally, the stability selection is utilised to analyze the temporal and spatial dynamic patterns between biomarkers. We prove that correlate information can better describe the brain structural changes in patients with NL, MCI and AD. 
Experiments shows that the effectiveness of the impact of AD progression on brain function
synergistic deteriorating biomarkers.

\section*{Acknowledgment}
This work was supported by the China Scholarship Council (No.202107030007), Engineering and Physical Sciences Research Council (EPSRC) Doctoral Training Partnership (EP/T517835/1) and Young Scientists Fund of the National Natural Science Foundation of China (Grant No.62301452).
\bibliographystyle{IEEEtran}
\bibliography{main}
\onecolumn
\appendix \label{appendix}
    \begin{table*}[h]
    \caption{THE STABILITY ROIS PAIRS AND THE CORRESPONDING NEUROANATOMY (ADAS-COG-TARGETED)}
    \label{THE STABILITY ROIS PAIRS AND THE CORRESPONDING NEUROANATOMY ADAS-COG-TARGETED}
    \resizebox{\textwidth}{!}{%
    \begin{tabular}{@{}ll@{}}
    \toprule
    RIO Pairs      & Defination                                                                                                                    \\ \midrule                                                                                       
    ST104SA-ST62CV & ['Surface Area of RightParsOpercularis $\Leftrightarrow$ Volume   (Cortical Parcellation) of LeftTransverseTemporal']                               \\
    ST109CV-ST13SA & ['Volume (Cortical Parcellation) of   RightPosteriorCingulate $\Leftrightarrow$ Surface Area of LeftBankssts']    \\
    ST109SA-ST12SV & ['Surface Area of RightPosteriorCingulate $\Leftrightarrow$ Volume (WM   Parcellation) of LeftAmygdala']          \\
    ST109SA-ST47SA & ['Surface Area of RightPosteriorCingulate $\Leftrightarrow$ Surface Area   of LeftParsTriangularis']              \\
    ST109TA-ST44SA & ['Cortical Thickness Average of   RightPosteriorCingulate $\Leftrightarrow$ Surface Area of LeftParahippocampal'] \\
    ST109TA-ST48SA & ['Cortical Thickness Average of   RightPosteriorCingulate $\Leftrightarrow$ Surface Area of LeftPericalcarine']   \\
    ST109TS-ST30SV & ['Cortical Thickness Standard Deviation of   RightPosteriorCingulate $\Leftrightarrow$ Volume (WM Parcellation) of   LeftInferiorLateralVentricle'] \\
    ST112SV-ST50SA & ['Volume (WM Parcellation) of RightPutamen $\Leftrightarrow$ Surface   Area of LeftPosteriorCingulate']           \\
    ST116TS-ST12SV & ['Cortical Thickness Standard Deviation of RightSuperiorParietal $\Leftrightarrow$ Volume   (WM Parcellation) of LeftAmygdala']                     \\
    ST118TS-ST62CV & ['Cortical Thickness Standard Deviation of   RightSupramarginal $\Leftrightarrow$ Volume (Cortical Parcellation) of   LeftTransverseTemporal']      \\
    ST129SA-ST13SA & ['Surface Area of LeftInsula $\Leftrightarrow$ Surface Area of   LeftBankssts']                                   \\
    ST129SA-ST25CV & ['Surface Area of LeftInsula $\Leftrightarrow$ Volume (Cortical   Parcellation) of LeftFrontalPole']              \\
    ST129SA-ST47SA & ['Surface Area of LeftInsula $\Leftrightarrow$ Surface Area of   LeftParsTriangularis']                           \\
    ST13SA-ST48SA  & ['Surface Area of LeftBankssts $\Leftrightarrow$ Surface Area of   LeftPericalcarine']                            \\
    ST15TA-ST48SA  & ['Cortical Thickness Average of   LeftCaudalMiddleFrontal $\Leftrightarrow$ Surface Area of LeftPericalcarine']   \\
    ST24SA-ST44SA  & ['Surface Area of LeftEntorhinal $\Leftrightarrow$ Surface Area of   LeftParahippocampal']                        \\
    ST24SA-ST71SV  & ['Surface Area of LeftEntorhinal $\Leftrightarrow$ Volume (WM   Parcellation) of RightAmygdala']                  \\
    ST24SA-ST82SA  & ['Surface Area of LeftEntorhinal $\Leftrightarrow$ Surface Area of   RightCuneus']                                \\
    ST24SA-ST98SA  & ['Surface Area of LeftEntorhinal $\Leftrightarrow$ Surface Area of   RightMedialOrbitofrontal']                   \\
    ST24TA-ST50SA  & ['Cortical Thickness Average of LeftEntorhinal $\Leftrightarrow$ Surface   Area of LeftPosteriorCingulate']       \\
    ST32TA-ST71SV  & ['Cortical Thickness Average of   LeftInferiorTemporal $\Leftrightarrow$ Volume (WM Parcellation) of RightAmygdala']                                \\
    ST34TA-ST50SA  & ['Cortical Thickness Average of   LeftIsthmusCingulate $\Leftrightarrow$ Surface Area of LeftPosteriorCingulate'] \\
    ST45TA-ST93SA  & ['Cortical Thickness Average of   LeftParsOpercularis $\Leftrightarrow$ Surface Area of RightIsthmusCingulate']   \\
    ST49TA-ST62CV  & ['Cortical Thickness Average of LeftPostcentral $\Leftrightarrow$ Volume   (Cortical Parcellation) of LeftTransverseTemporal']                      \\
    ST51SA-ST62CV  & ['Surface Area of LeftPrecentral $\Leftrightarrow$ Volume (Cortical   Parcellation) of LeftTransverseTemporal']   \\
    ST51SA-ST71SV  & ['Surface Area of LeftPrecentral $\Leftrightarrow$ Volume (WM   Parcellation) of RightAmygdala']                  \\
    ST56TS-ST93SA  & ['Cortical Thickness Standard Deviation of   LeftSuperiorFrontal $\Leftrightarrow$ Surface Area of RightIsthmusCingulate']                          \\
    ST73TS-ST93SA  & ['Cortical Thickness Standard Deviation of   RightCaudalAnteriorCingulate $\Leftrightarrow$ Surface Area of   RightIsthmusCingulate']               \\
    ST90TA-ST93SA  & ['Cortical Thickness Average of   RightInferiorParietal $\Leftrightarrow$ Surface Area of RightIsthmusCingulate'] \\ \bottomrule
    \end{tabular}%
    }
    \end{table*}

    \begin{table*}[]
    \caption{THE STABILITY ROIS PAIRS AND THE CORRESPONDING NEUROANATOMY (MMSE-TARGETED)}
    \label{THE STABILITY ROIS PAIRS AND THE CORRESPONDING NEUROANATOMY}
    \resizebox{\textwidth}{!}{%
    \begin{tabular}{@{}ll@{}}
    \toprule
    RIO Pairs      & Defination                                                                                                                    \\ \midrule
    ST101SV-ST13SA & ['Volume (WM Parcellation) of   RightPallidum $\Leftrightarrow$ Surface Area of LeftBankssts']                                              \\
    ST102SA-ST46CV & ['Surface Area of RightParacentral $\Leftrightarrow$ Volume   (Cortical Parcellation) of LeftParsOrbitalis']                                \\
    ST103TA-ST44CV & ['Cortical Thickness Average of   RightParahippocampal $\Leftrightarrow$ Volume (Cortical Parcellation) of   LeftParahippocampal']          \\
    ST104CV-ST46SA & ['Volume (Cortical Parcellation) of   RightParsOpercularis $\Leftrightarrow$ Surface Area of LeftParsOrbitalis']                            \\
    ST105TA-ST46CV & ['Cortical Thickness Average of   RightParsOrbitalis $\Leftrightarrow$ Volume (Cortical Parcellation) of   LeftParsOrbitalis']              \\
    ST107CV-ST70SV & ['Volume (Cortical Parcellation) of   RightPericalcarine $\Leftrightarrow$ Volume (WM Parcellation) of RightAccumbensArea']                 \\
    ST107TA-ST93CV & ['Cortical Thickness Average of   RightPericalcarine $\Leftrightarrow$ Volume (Cortical Parcellation) of   RightIsthmusCingulate']          \\
    ST108TS-ST60TA & ['Cortical Thickness Standard Deviation of   RightPostcentral $\Leftrightarrow$ Cortical Thickness Average of LeftTemporalPole']            \\
    ST109SA-ST12SV & ['Surface Area of   RightPosteriorCingulate $\Leftrightarrow$ Volume (WM Parcellation) of LeftAmygdala']                                    \\
    ST109SA-ST93CV & ['Surface Area of   RightPosteriorCingulate $\Leftrightarrow$ Volume (Cortical Parcellation) of   RightIsthmusCingulate']                   \\
    ST109TS-ST14CV & ['Cortical Thickness Standard Deviation of   RightPosteriorCingulate $\Leftrightarrow$ Volume (Cortical Parcellation) of   LeftCaudalAnteriorCingulate']       \\
    ST109TS-ST48TA & ['Cortical Thickness Standard Deviation of   RightPosteriorCingulate $\Leftrightarrow$ Cortical Thickness Average of   LeftPericalcarine']  \\
    ST109TS-ST60CV & ['Cortical Thickness Standard Deviation of   RightPosteriorCingulate $\Leftrightarrow$ Volume (Cortical Parcellation) of   LeftTemporalPole']                  \\
    ST110SA-ST98SA & ['Surface Area of RightPrecentral $\Leftrightarrow$ Surface   Area of RightMedialOrbitofrontal']                                            \\
    ST110TA-ST44SA & ['Cortical Thickness Average of   RightPrecentral $\Leftrightarrow$ Surface Area of LeftParahippocampal']                                   \\
    ST110TA-ST74SA & ['Cortical Thickness Average of   RightPrecentral $\Leftrightarrow$ Surface Area of RightCaudalMiddleFrontal']                              \\
    ST110TS-ST25CV & ['Cortical Thickness Standard Deviation of   RightPrecentral $\Leftrightarrow$ Volume (Cortical Parcellation) of LeftFrontalPole']          \\
    ST111CV-ST46SA & ['Volume (Cortical Parcellation) of   RightPrecuneus $\Leftrightarrow$ Surface Area of LeftParsOrbitalis']                                  \\
    ST111TA-ST25CV & ['Cortical Thickness Average of   RightPrecuneus $\Leftrightarrow$ Volume (Cortical Parcellation) of LeftFrontalPole']                      \\
    ST113SA-ST39SA & ['Surface Area of   RightRostralAnteriorCingulate $\Leftrightarrow$ Surface Area of   LeftMedialOrbitofrontal']                             \\
    ST113SA-ST43SA & ['Surface Area of   RightRostralAnteriorCingulate $\Leftrightarrow$ Surface Area of LeftParacentral']                                       \\
    ST113SA-ST74SA & ['Surface Area of   RightRostralAnteriorCingulate $\Leftrightarrow$ Surface Area of   RightCaudalMiddleFrontal']                            \\
    ST114TA-ST43SA & ['Cortical Thickness Average of   RightRostralMiddleFrontal $\Leftrightarrow$ Surface Area of LeftParacentral']                             \\
    ST114TA-ST93CV & ['Cortical Thickness Average of   RightRostralMiddleFrontal $\Leftrightarrow$ Volume (Cortical Parcellation) of   RightIsthmusCingulate']   \\
    ST115CV-ST48CV & ['Volume (Cortical Parcellation) of   RightSuperiorFrontal $\Leftrightarrow$ Volume (Cortical Parcellation) of   LeftPericalcarine']        \\
    ST115TS-ST42SV & ['Cortical Thickness Standard Deviation of   RightSuperiorFrontal $\Leftrightarrow$ Volume (WM Parcellation) of LeftPallidum']              \\
    ST115TS-ST72CV & ['Cortical Thickness Standard Deviation of   RightSuperiorFrontal $\Leftrightarrow$ Volume (Cortical Parcellation) of   RightBankssts']     \\
    ST116TS-ST25CV & ['Cortical Thickness Standard Deviation of   RightSuperiorParietal $\Leftrightarrow$ Volume (Cortical Parcellation) of   LeftFrontalPole']  \\
    ST116TS-ST93CV & ['Cortical Thickness Standard Deviation of   RightSuperiorParietal $\Leftrightarrow$ Volume (Cortical Parcellation) of   RightIsthmusCingulate']               \\
    ST119TA-ST72CV & ['Cortical Thickness Average of   RightTemporalPole $\Leftrightarrow$ Volume (Cortical Parcellation) of RightBankssts']                     \\
    ST120SV-ST46SA & ['Volume (WM Parcellation) of   RightThalamus $\Leftrightarrow$ Surface Area of LeftParsOrbitalis']                                         \\
    ST123SA-ST83CV & ['Surface Area of RightUnknown $\Leftrightarrow$ Volume   (Cortical Parcellation) of RightEntorhinal']                                      \\
    ST124SV-ST71SV & ['Volume (WM Parcellation) of   RightVentralDC $\Leftrightarrow$ Volume (WM Parcellation) of RightAmygdala']                                \\
    ST129CV-ST24CV & ['Volume (Cortical Parcellation) of   LeftInsula $\Leftrightarrow$ Volume (Cortical Parcellation) of LeftEntorhinal']                       \\
    ST129CV-ST50SA & ['Volume (Cortical Parcellation) of   LeftInsula $\Leftrightarrow$ Surface Area of LeftPosteriorCingulate']                                 \\
    ST129CV-ST72CV & ['Volume (Cortical Parcellation) of   LeftInsula $\Leftrightarrow$ Volume (Cortical Parcellation) of RightBankssts']                        \\
    ST129CV-ST84SA & ['Volume (Cortical Parcellation) of   LeftInsula $\Leftrightarrow$ Surface Area of RightFrontalPole']                                       \\
    ST129SA-ST14CV & ['Surface Area of LeftInsula $\Leftrightarrow$ Volume   (Cortical Parcellation) of LeftCaudalAnteriorCingulate']                            \\
    ST14TA-ST46SA  & ['Cortical Thickness Average of   LeftCaudalAnteriorCingulate $\Leftrightarrow$ Surface Area of LeftParsOrbitalis']                         \\
    ST15SA-ST24CV  & ['Surface Area of LeftCaudalMiddleFrontal $\Leftrightarrow$ Volume   (Cortical Parcellation) of LeftEntorhinal']                            \\
    ST15SA-ST30SV  & ['Surface Area of   LeftCaudalMiddleFrontal $\Leftrightarrow$ Volume (WM Parcellation) of   LeftInferiorLateralVentricle']                  \\
    ST15SA-ST9SV   & ['Surface Area of LeftCaudalMiddleFrontal $\Leftrightarrow$ Volume   (WM Parcellation) of FourthVentricle']                                 \\
    ST19SV-ST43SA  & ['Volume (WM Parcellation) of   LeftCerebralCortex $\Leftrightarrow$ Surface Area of LeftParacentral']                                      \\
    ST19SV-ST54CV  & ['Volume (WM Parcellation) of   LeftCerebralCortex $\Leftrightarrow$ Volume (Cortical Parcellation) of   LeftRostralAnteriorCingulate']     \\
    ST19SV-ST55TA  & ['Volume (WM Parcellation) of   LeftCerebralCortex $\Leftrightarrow$ Cortical Thickness Average of   LeftRostralMiddleFrontal']             \\
    ST24SA-ST34CV  & ['Surface Area of LeftEntorhinal $\Leftrightarrow$ Volume   (Cortical Parcellation) of LeftIsthmusCingulate']                               \\
    ST24TA-ST30SV  & ['Cortical Thickness Average of   LeftEntorhinal $\Leftrightarrow$ Volume (WM Parcellation) of   LeftInferiorLateralVentricle']             \\
    ST31CV-ST98SA  & ['Volume (Cortical Parcellation) of   LeftInferiorParietal $\Leftrightarrow$ Surface Area of RightMedialOrbitofrontal']                     \\
    ST31SA-ST46CV  & ['Surface Area of   LeftInferiorParietal $\Leftrightarrow$ Volume (Cortical Parcellation) of   LeftParsOrbitalis']                          \\
    ST32TA-ST62CV  & ['Cortical Thickness Average of   LeftInferiorTemporal $\Leftrightarrow$ Volume (Cortical Parcellation) of   LeftTransverseTemporal']       \\
    ST34CV-ST44CV  & ['Volume (Cortical Parcellation) of   LeftIsthmusCingulate $\Leftrightarrow$ Volume (Cortical Parcellation) of   LeftParahippocampal']      \\
    ST34CV-ST93CV  & ['Volume (Cortical Parcellation) of LeftIsthmusCingulate $\Leftrightarrow$ Volume   (Cortical Parcellation) of RightIsthmusCingulate']      \\
    ST35SA-ST47SA  & ['Surface Area of   LeftLateralOccipital $\Leftrightarrow$ Surface Area of LeftParsTriangularis']                                           \\
    ST39SA-ST43SA  & ['Surface Area of   LeftMedialOrbitofrontal $\Leftrightarrow$ Surface Area of LeftParacentral']                                             \\
    ST39SA-ST70SV  & ['Surface Area of   LeftMedialOrbitofrontal $\Leftrightarrow$ Volume (WM Parcellation) of   RightAccumbensArea']                            \\
    ST45TA-ST93SA  & ['Cortical Thickness Average of   LeftParsOpercularis $\Leftrightarrow$ Surface Area of RightIsthmusCingulate']                             \\
    ST47CV-ST72CV  & ['Volume (Cortical Parcellation) of   LeftParsTriangularis $\Leftrightarrow$ Volume (Cortical Parcellation) of   RightBankssts']            \\
    ST47CV-ST73CV  & ['Volume (Cortical Parcellation) of   LeftParsTriangularis $\Leftrightarrow$ Volume (Cortical Parcellation) of   RightCaudalAnteriorCingulate']                \\
    ST48TS-ST74SA  & ['Cortical Thickness Standard Deviation of   LeftPericalcarine $\Leftrightarrow$ Surface Area of RightCaudalMiddleFrontal']                 \\
    ST51CV-ST60CV  & ['Volume (Cortical Parcellation) of   LeftPrecentral $\Leftrightarrow$ Volume (Cortical Parcellation) of LeftTemporalPole']                 \\
    ST53SV-ST5SV   & ['Volume (WM Parcellation) of   LeftPutamen $\Leftrightarrow$ Volume (WM Parcellation) of   CorpusCallosumMidPosterior']                    \\
    ST53SV-ST7SV   & ['Volume (WM Parcellation) of   LeftPutamen $\Leftrightarrow$ Volume (WM Parcellation) of Csf']                                             \\
    ST54TS-ST69SV  & ['Cortical Thickness Standard Deviation of   LeftRostralAnteriorCingulate $\Leftrightarrow$ Volume (WM Parcellation) of   OpticChiasm']     \\
    ST55TA-ST83CV  & ['Cortical Thickness Average of   LeftRostralMiddleFrontal $\Leftrightarrow$ Volume (Cortical Parcellation) of   RightEntorhinal']          \\
    ST55TS-ST73CV  & ['Cortical Thickness Standard Deviation of   LeftRostralMiddleFrontal $\Leftrightarrow$ Volume (Cortical Parcellation) of   RightCaudalAnteriorCingulate']     \\
    ST55TS-ST95TS  & ['Cortical Thickness Standard Deviation of   LeftRostralMiddleFrontal $\Leftrightarrow$ Cortical Thickness Standard Deviation of   RightLateralOrbitofrontal'] \\
    ST56TS-ST84SA  & ['Cortical Thickness Standard Deviation of   LeftSuperiorFrontal $\Leftrightarrow$ Surface Area of RightFrontalPole']                       \\
    ST59SA-ST62SA  & ['Surface Area of LeftSupramarginal $\Leftrightarrow$ Surface   Area of LeftTransverseTemporal']                                            \\
    ST60SA-ST70SV  & ['Surface Area of LeftTemporalPole $\Leftrightarrow$ Volume   (WM Parcellation) of RightAccumbensArea']                                     \\
    ST62TS-ST83CV  & ['Cortical Thickness Standard Deviation of   LeftTransverseTemporal $\Leftrightarrow$ Volume (Cortical Parcellation) of   RightEntorhinal'] \\
    ST73CV-ST93CV  & ['Volume (Cortical Parcellation) of   RightCaudalAnteriorCingulate $\Leftrightarrow$ Volume (Cortical Parcellation) of   RightIsthmusCingulate']               \\ \bottomrule
    \end{tabular}%
    }
    \end{table*}

\onecolumn
\end{document}